\documentclass[lettersize,journal]{IEEEtran}
\usepackage{amsmath,amsfonts}
\usepackage{algorithmic}
\usepackage{algorithm}
\usepackage{array}
\usepackage[caption=false,font=normalsize,labelfont=sf,textfont=sf]{subfig}
\usepackage{textcomp}
\usepackage{stfloats}
\usepackage{url}
\usepackage{verbatim}
\usepackage{graphicx}
\usepackage{cite}
\usepackage{amsmath} 
\usepackage{indentfirst} 
\usepackage{gensymb}
\usepackage{graphicx} 
\usepackage{amsmath} 
\usepackage{xcolor} 
\usepackage{fancyhdr} 

\usepackage{booktabs}
\usepackage{multirow}

\begin{document}
\title{Effects of Assistance Delay on Joint Mechanics and Energetics in Biological Torque Control of a Hip Exoskeleton}

\author{Jimin An, Ryan Lee, Jingshu Peng, Eni Halilaj, and Inseung Kang
\thanks{J. An, R. Lee, J. Peng, E. Halilaj, and I. Kang are with the Department of Mechanical Engineering, Carnegie Mellon University, Pittsburgh, PA, 15213 USA (Corresponding author: Jimin An, email: {\tt\footnotesize jiminan@cmu.edu}).}}




\maketitle

\begin{abstract}
Biological torque control directly maps an estimated human joint moment to exoskeleton assistance, providing a task-agnostic strategy for supporting diverse locomotor activities. However, it remains unclear whether a fixed state-to-torque mapping provides effective assistance across biomechanically distinct tasks. We examined how assistance delay affected hip exoskeleton performance during level-ground (LG), ramp-ascent (RA), and ramp-descent (RD) walking. Eight participants completed a zero-torque baseline condition and five active assistance conditions with delays ranging from 40 to 320~ms. Across tasks and active delays, assistance reduced net metabolic rate by 5.24\%, positive biological hip joint work by 5.86\%, and total lower-limb positive joint work by 1.68\% (all \(p<0.05\)). Assistance delay affected both joint-work outcomes (both \(p<0.001\)) but not net metabolic rate. Mechanical unloading generally decreased with increasing delay, whereas metabolic benefits remained comparatively stable. Relative to the zero-torque condition, net metabolic rate decreased by 9.75\% during LG and 7.20\% during RA but increased by 1.23\% during RD. We did not detect task-dependent differences in the delay response. Our findings indicate that biological torque mappings should be evaluated based on the target outcome and mechanical role of the assisted joint, and that predominantly positive-power assistance may not generalize to negative-work-dominant locomotion without modification.
\end{abstract}

\begin{IEEEkeywords}
Robotic Hip Exoskeleton, Biological Torque Control, Metabolic Cost, Joint Mechanics
\end{IEEEkeywords}

\section{Introduction}
\IEEEPARstart{L}{ower-limb} exoskeletons can reduce the physical effort required for locomotion and support mobility across a range of applications \cite{YoungSOTA, YoungExpansion}. However, translating these benefits beyond controlled laboratory conditions remains challenging, in large part because exoskeleton assistance must adapt to changes in the user and the locomotor task. Conventional controllers commonly use a cascaded architecture that first estimates locomotor states, such as locomotion mode \cite{Kang_TBME_2022, Qian_RAL_2022} or gait phase \cite{Medrano_TRO_2023, SamsungHipMetabolic, GPE1}, and then uses these estimates to generate an assistance profile. Although this approach can provide effective assistance during steady-state locomotion, its performance may degrade during task transitions and non-cyclic movements. Moreover, it is impractical to design and tune assistance profiles across the full range of user states encountered during real-world locomotion.

To address these limitations, recent exoskeleton control research has focused on using the user’s physiological state in real time. Such physiological state information may enable task-agnostic assistance without requiring explicit recognition of external locomotor states. Biological torque control implements this approach by estimating the user’s joint moment in real time and mapping the estimate directly to exoskeleton torque \cite{Molinaro_TMRB_2022, Molinaro_Sci_Robot}. Because the estimated joint moment and exoskeleton command are expressed in the same physical domain, this framework can generate assistance without explicitly identifying the locomotor state. Biological torque-based hip exoskeleton control has demonstrated effectiveness across steady-state, transient, cyclic, and non-cyclic movements, and has reduced metabolic cost and positive joint work during level-ground (LG) and ramp-ascent (RA) walking \cite{Molinaro_Sci_Robot}.

Although the current biological torque controller provides task-agnostic assistance, it remains unclear whether its state-to-torque mapping is optimal across tasks. While the physiological-state estimator can estimate biological joint moments across diverse movements, the subsequent mapping to exoskeleton torque is governed by fixed scaling and delay parameters \cite{Molinaro_Nature}. This fixed mapping may be limiting because the magnitude and timing of assistance strongly influence its benefit \cite{YoungMagnitude, YoungTiming}. Previous optimization studies using parameterized assistance profiles have shown that optimal assistance can vary across locomotor tasks \cite{Ding2016HipTiming, ConorWalkRun, Stair, FranksIncline}. Importantly, the effect of these mapping parameters may also depend on how assistance benefit is quantified. Joint-work outcomes more directly reflect mechanical unloading at the assisted joint, whereas metabolic cost integrates neuromuscular and physiological responses across the body. Differences in optimal assistance are consistent with the distinct mechanical demands of each task. RA walking requires greater positive hip mechanical work and altered hip power timing compared with LG walking, whereas ramp-descent (RD) walking involves substantial negative mechanical work associated with energy absorption \cite{FranksIncline, UpDownhillOpt, UphillDownhillJointWork}. Thus, although a fixed state-to-torque mapping may enable task-generalizable assistance, different locomotor tasks may benefit from different mapping parameters.

Existing studies provide mixed evidence regarding whether the biological torque mapping should be task-specific. An initial pilot study found that favorable delay ranges were similar for LG and RA walking, but its small sample size precluded a systematic comparison across tasks \cite{Molinaro_Sci_Robot}. A subsequent human-in-the-loop optimization study found clear benefits of user-specific optimization but no significant advantage of task-specific over task-agnostic assistance across the evaluated walking conditions \cite{Justine}. Separately, an ankle exoskeleton study showed that varying assistance delay and magnitude produced different effects on muscle activation and biological joint moment even within a single locomotor task \cite{JuanjuanAnkleExo}.
Taken together, the available evidence is insufficient to determine whether biomechanically distinct locomotor tasks require different state-to-torque mappings.

In this study, we investigated how assistance delay affects the biomechanical, metabolic, and subjective outcomes of hip exoskeleton assistance under biological torque control during LG, RA, and RD walking. First, we hypothesized that the effect of assistance delay would differ across net metabolic rate, positive biological hip joint work, and total lower-limb positive joint work (H1). Second, given the distinct mechanical demands of the three locomotor tasks, we hypothesized that the effect of assistance delay would depend on the locomotor task (H2). We additionally explored whether subjective ratings varied systematically with assistance delay and locomotor task.

\section{Robotic Hip Exoskeleton}
\subsection{Hardware Design}
\begin{figure}
    \includegraphics[width=1\columnwidth]{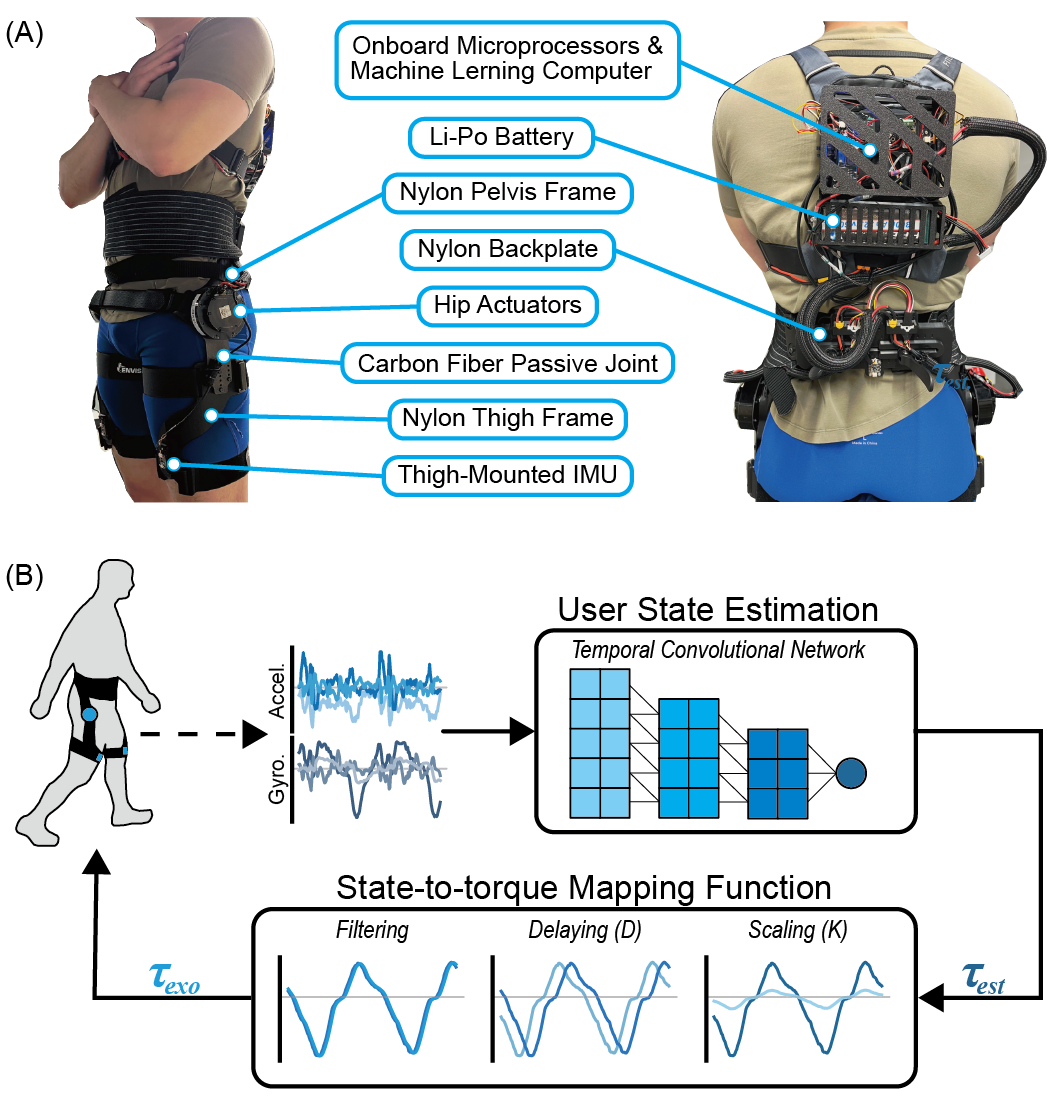}
    \caption{
       \textbf{Bilateral robotic hip exoskeleton and biological torque controller scheme.}
       (A) The exoskeleton provides bilateral hip flexion and extension assistance during locomotion. Actuators are located at the user's hip joints, and thigh-mounted IMUs provide kinematic information. The user interfaces are adjustable to accommodate varying body sizes. (B) The controller receives IMU inputs and estimates the biological hip joint moment independently for each limb. A predefined state-to-torque mapping function then scales, delays, and filters each estimate to generate the exoskeleton torque applied to the user.
       }
    \label{CombFig}
\end{figure}

We developed a bilateral robotic hip exoskeleton to deliver assistive torque during locomotion (Fig.~\ref{CombFig}A). The exoskeleton was designed to provide up to 18~Nm of hip torque across a sagittal-plane range of motion from 30$^\circ$ of extension to 100$^\circ$ of flexion. A compliant carbon-fiber passive joint permitted $\pm5^\circ$ of hip abduction and adduction in the frontal plane. The device weighed 4.3~kg and included two actuators (AK80-9; CubeMars, Nanchang, China), 3D-printed nylon user interfaces, body straps, and onboard electronics. The exoskeleton was secured to the user with adjustable straps at the thigh, waist, and torso, which accommodated diverse body sizes. The embedded encoder on each actuator provided angle and angular velocity measurements. An inertial measurement unit (IMU) was mounted distally on each thigh frame to record limb kinematics (ICM-20948; Adafruit Industries, Brooklyn, NY, USA). The onboard electronics backpack housed a 22.2~V, 3300~mAh lithium-polymer battery and an embedded computing platform (Jetson Orin Nano Developer Kit; NVIDIA, Santa Clara, CA, USA) for sensor data acquisition and actuator control.

\subsection{Biological Torque Controller}
The biological torque controller estimates the instantaneous hip joint moment using a deep-learning model and applies a reshaped version of the estimate as the exoskeleton torque (Fig.~\ref{CombFig}B). The exoskeleton torque applied at time $t$, $\tau_{\mathrm{exo}}(t)$, was computed as
\begin{equation}
\label{eq:mapping}
    \tau_{\mathrm{exo}}(t)=\mathcal{F}\left\{K \cdot \tau_{\mathrm{est}}(t-D)\right\},
\end{equation}
where $\tau_{\mathrm{est}}$ denotes the estimated biological hip joint moment, $\mathcal{F}$ denotes the filtering operation, and $K$ and $D$ denote the scaling factor and delay parameter, respectively. In this study, $K$ was fixed at 0.2 for all assisted conditions. The commanded exoskeleton torque was capped at 18~Nm to stay within the actuator torque limit.

We trained a temporal convolutional network (TCN) to estimate instantaneous hip joint moments, following the architecture and hyperparameter settings reported in a previous study \cite{Molinaro_Sci_Robot}. The model used six input channels from a single thigh-mounted IMU and a 950~ms input window, with data sampled at 100~Hz to match the experimental setup. The objective of this study was to evaluate the state-to-torque mapping rather than subject-independent or task-generalizable estimation. Accordingly, we trained a separate biological joint moment estimator for each locomotor task using exoskeleton-assisted walking data collected from 10 healthy participants. The task-specific models were validated on data collected during the main experiment and achieved root mean square error (RMSE) and coefficient of determination ($R^2$) values (Table~\ref{tab:rmse_r2_avg}) comparable to those reported previously \cite{Molinaro_Sci_Robot}.


\begin{table}[t]
\setlength{\tabcolsep}{3pt}
\renewcommand{\arraystretch}{0.8}
\centering
\small
\caption{Real-time biological torque estimator performance}
\label{tab:rmse_r2_avg}
\begin{tabular}{lccc}
\toprule
\toprule
\textbf{Task} & \textbf{RMSE (Nm/kg)} & \textbf{R$^{2}$} \\
\midrule
LG & 0.07$\pm$0.01 & 0.95$\pm$0.02 \\
RA & 0.12$\pm$0.04 & 0.90$\pm$0.04 \\
RD & 0.08$\pm$0.02 & 0.86$\pm$0.06 \\
\midrule
\textbf{Average} & \textbf{0.09$\pm$0.03} & \textbf{0.90$\pm$0.06} \\
\bottomrule
\bottomrule
\end{tabular}
\end{table}


\section{Methods}
\subsection{Experiment Design}

\begin{figure}[!t]
    \includegraphics[width=1\columnwidth]{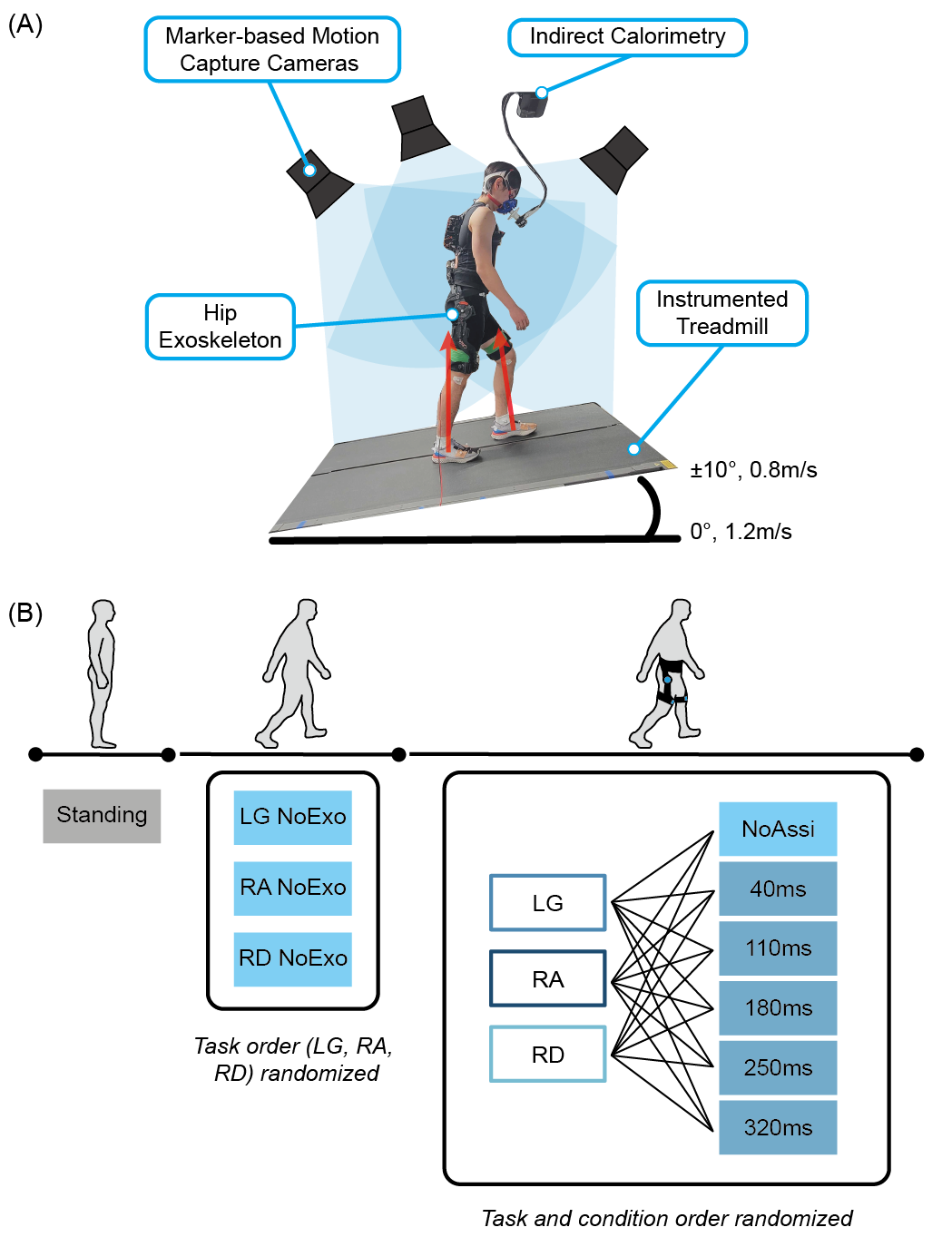}
    \caption{
       \textbf{Experimental setup and protocol.} (A) Participants performed level-ground (LG), ramp-ascent (RA), and ramp-descent (RD) walking on an instrumented treadmill while metabolic rate was measured by indirect calorimetry and kinematics were recorded with reflective motion-capture markers. Red arrows indicate the ground-reaction forces measured by the instrumented treadmill. (B) Experimental sequence comprising standing, NoExo, NoAssi, and five active-delay conditions across the three locomotor tasks. Task order was randomized for both the NoExo and exoskeleton trials, and the order of the active-delay conditions was independently randomized to minimize potential order effects.
    }
    \label{ExperimentProtocol}
\end{figure}

The study protocol was approved by the Institutional Review Board of Carnegie Mellon University on April 1, 2025 (Protocol STUDY2025\_00000002), and all participants provided written informed consent before the experiment. Eight healthy adults were recruited for this study (seven males and one female; age: $26.4 \pm 2.9$ years; body mass: $71.8 \pm 9.8$~kg; height: $1.77 \pm 0.07$~m). Participants walked on an instrumented treadmill (FIT5; Bertec, Columbus, OH, USA) for 6~min under each experimental condition. Walking speed was set to 1.2~m/s for LG walking and 0.8~m/s for RA and RD walking, which were performed at inclines of $+10^{\circ}$ and $-10^{\circ}$, respectively (Fig.~\ref{ExperimentProtocol}A). Different walking speeds were selected to maintain comparable physiological demands across tasks and to minimize the contribution of anaerobic metabolism.

Seven experimental conditions were tested for each locomotor task: walking without the exoskeleton (NoExo), walking with the exoskeleton under zero commanded torque (NoAssi), and five assisted conditions with delays ranging from 40 to 320~ms in 70~ms increments. The lower bound was determined by the fixed 40~ms system latency, consisting of approximately 10~ms for estimator inference and 30~ms for signal filtering. The upper bound was determined from pilot testing, in which participants reported discomfort at delays greater than 320~ms during LG walking. All reported delay values represent the total effective delay, including both the programmed delay and the fixed system latency.

Each session began with a standing trial to measure resting metabolic rate. Participants then completed the NoExo condition for each of the three locomotor tasks to establish task-specific metabolic baselines (Fig.~\ref{ExperimentProtocol}B). Each of the remaining exoskeleton conditions, including NoAssi, consisted of 6~min of walking followed by 4~min of seated rest. A 20~min rest period was provided between locomotor tasks to reduce participant fatigue. To minimize potential order effects, both the task order and the order of the exoskeleton conditions within each task were randomized.

The effects of exoskeleton assistance were evaluated using metabolic, biomechanical, and subjective outcome measures. Metabolic rate was measured during each trial using a portable indirect calorimetry system (K5; COSMED, Rome, Italy). Joint kinematics and kinetics were computed from marker trajectories, recorded with a marker-based motion-capture system (Vero v2.2; Vicon, Oxford, UK), and from ground-reaction forces, measured synchronously by the instrumented treadmill. After each assisted trial, participants rated the assistance on a five-point preference scale, considering both perceived comfort and assistance benefit. Ratings were made independently for each condition without direct comparison with the other assistance conditions.

\subsection{Data Processing and Analysis}
\subsubsection{Biomechanical Outcomes}
Raw motion-capture and ground-reaction force data were processed in OpenSim \cite{OpenSim} using the gait2392 musculoskeletal model to obtain lower-limb joint kinematics and kinetics. Marker trajectories and ground-reaction force data were low-pass filtered using zero-lag, fifth-order Butterworth filters with cutoff frequencies of 6 and 20~Hz, respectively. For exoskeleton trials, the subject-specific model was first scaled using the NoExo static trial because the greater trochanter marker was relocated to the exoskeleton actuator when the device was worn. The OpenSim MarkerPlacer tool was then rerun using the static trial collected with the exoskeleton. To account for the exoskeleton mass, 1.3~kg was added to the torso, 2.0~kg to the pelvis, and 0.5~kg to each femur segment of the scaled model. Inverse kinematics and inverse dynamics were subsequently performed, and the resulting joint-angle and joint-moment trajectories were low-pass filtered at 4~Hz.

During assisted conditions, the sagittal-plane hip joint moment obtained from inverse dynamics represents the combined contributions of the user and the exoskeleton. The biological hip joint moment was therefore computed as
\begin{equation}
\label{biologicaltorque}
\tau_{\mathrm{bio}}(t) = \tau_{\mathrm{ID}}(t) - \tau_{\mathrm{exo}}(t),
\end{equation}
where $\tau_{\mathrm{ID}}$ denotes the net hip joint moment from inverse dynamics and $\tau_{\mathrm{exo}}$ denotes the recorded exoskeleton torque. Biological joint power was computed from the biological joint moment and joint angular velocity. Positive joint work was obtained by integrating the positive portion of joint power over each gait cycle, and total lower-limb positive joint work was calculated as the sum of positive hip, knee, and ankle joint work. For each trial, 100 gait cycles were randomly selected from the final 2~min for analysis. For one participant, the recorded exoskeleton torque was unavailable for the 250~ms delay condition during LG walking. For this trial, exoskeleton torque was reconstructed from the motion-capture-derived hip joint moments and used in the biomechanical analysis.

\subsubsection{Metabolic Rate}
Breath-by-breath respiratory data from the last 2~min of each trial were used to compute gross metabolic rate normalized by body mass \cite{Brockway}:
\begin{equation}
\label{brocakway}
P_{\mathrm{met}} = \frac{0.278 \times \dot{V}O_{2} + 0.075 \times \dot{V}CO_{2}}{m}
\end{equation}
where $\dot{V}O_2$ and $\dot{V}CO_2$ denote the rates of oxygen consumption and carbon dioxide production (mL/min), respectively, and $m$ denotes body mass (kg). The resting metabolic rate measured during the standing trial was subtracted from gross metabolic rate to obtain net metabolic rate, which was used for all subsequent analyses. For the metabolic and biomechanical outcomes, values from the active assistance conditions were expressed as percent changes relative to the task-specific NoAssi condition. For each participant and outcome, an overall assistance response was calculated by averaging these percent changes across the five active delay conditions and three locomotor tasks.

\begin{figure*}[!t]
    \includegraphics[width=1\textwidth]{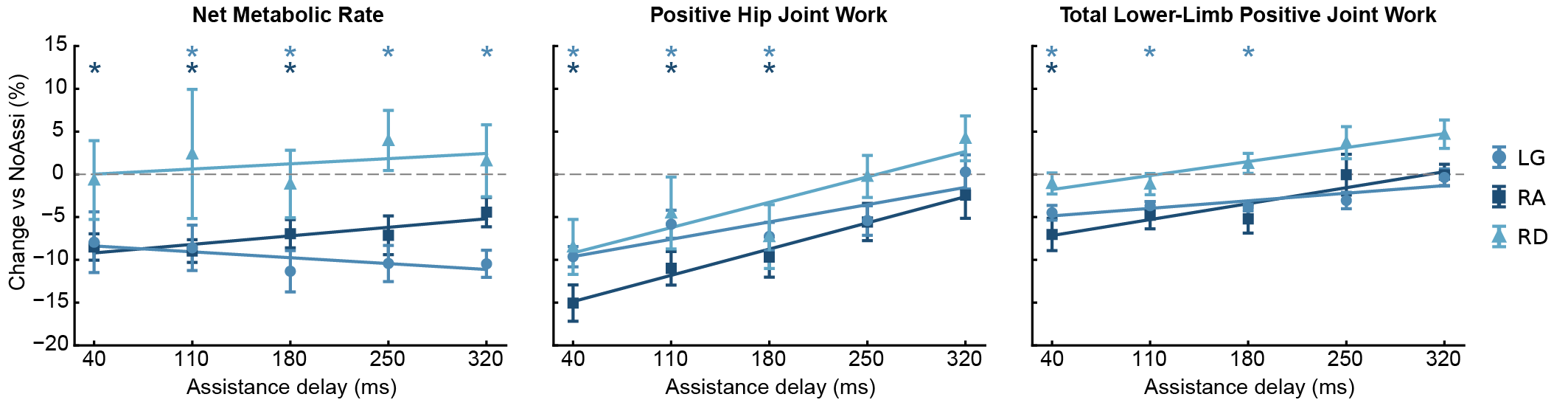}
    \caption{
       \textbf{Assistance-delay responses in metabolic and positive joint-work outcomes.} Panels show net metabolic rate, positive biological hip joint work, and total lower-limb positive joint work. Markers and error bars represent the mean percent change \(\pm\) SEM from the task-matched NoAssi condition across eight participants. Negative values indicate reductions relative to NoAssi, and the dashed horizontal line denotes no change. Circles, squares, and triangles represent level-ground (LG), ramp-ascent (RA), and ramp-descent (RD) walking, respectively. 
       Solid lines represent task-specific fixed-effect predictions from exploratory linear mixed-effects models that included Delay (continuous, scaled per 70-ms increment), Task, and their interaction as fixed effects and participant as a random intercept. Color-matched asterisks mark active-delay conditions that differed from the task-matched NoAssi condition (paired \(t\)-tests, \(p_{\mathrm{adj}}<0.05\), Bonferroni-corrected across the five delay comparisons within each task and outcome).
    }
    \label{DelayTaskEffect}
\end{figure*}
       
\subsection{Statistical Analysis}
To test our hypotheses, we performed a three-way repeated-measures analysis of variance (ANOVA) on the NoAssi-relative percent changes, with Outcome, Task, and Delay as within-subject factors. The Outcome \(\times\) Delay interaction tested H1 by determining whether delay-response patterns differed across net metabolic rate, positive biological hip joint work, and total lower-limb positive joint work.
The Task \(\times\) Delay interaction tested H2 by determining whether delay-response patterns depended on the locomotor task. The Outcome \(\times\) Task \(\times\) Delay interaction assessed whether this task dependence differed across outcome metrics. To characterize the response of each outcome, we subsequently conducted separate two-way repeated-measures ANOVAs with Task and Delay as within-subject factors.
Significant Delay or Task main effects were followed by pairwise comparisons using two-sided paired \(t\)-tests on participant-level marginal means, with Bonferroni correction applied separately within each outcome and factor.

To complement the ANOVAs, which treated delay as a categorical factor, we fitted exploratory linear mixed-effects models to characterize delay-response trends for each objective outcome. Each model included Delay, Task, and their interaction as fixed effects, with Delay treated as a continuous predictor that was mean-centered and scaled per 70~ms increment. Subject was included as a random intercept, and the models were fitted using maximum likelihood. Likelihood-ratio tests of nested models were used to evaluate whether an overall linear delay-response trend was present and whether its slope differed across tasks.

For each objective outcome, the overall effect of active assistance was evaluated with a two-sided one-sample \(t\)-test comparing the overall assistance response with zero. Separately, to identify active-delay conditions that differed from the task-specific NoAssi reference, each active-delay condition was compared with NoAssi using two-sided paired \(t\)-tests on the raw outcome values. Bonferroni correction was applied across the five delay comparisons within each task and outcome. Finally, subjective ratings were analyzed in an exploratory two-way repeated-measures ANOVA with Task and Delay as within-subject factors. Significant effects were followed by Bonferroni-corrected pairwise comparisons.

Statistical significance was defined as \(p<0.05\), and Bonferroni-adjusted \(p\)-values (\(p_{\mathrm{adj}}\)) are reported where applicable. To assess the sensitivity of the categorical analyses to task-dependent and subjective effects, we estimated model-conditional post hoc power and the total sample sizes required to achieve 80\% power. These estimates were obtained from 20000 parametric simulations based on the observed effect patterns and within-subject covariance structure.

\begin{figure}[!t]
    \includegraphics[width=1\columnwidth]{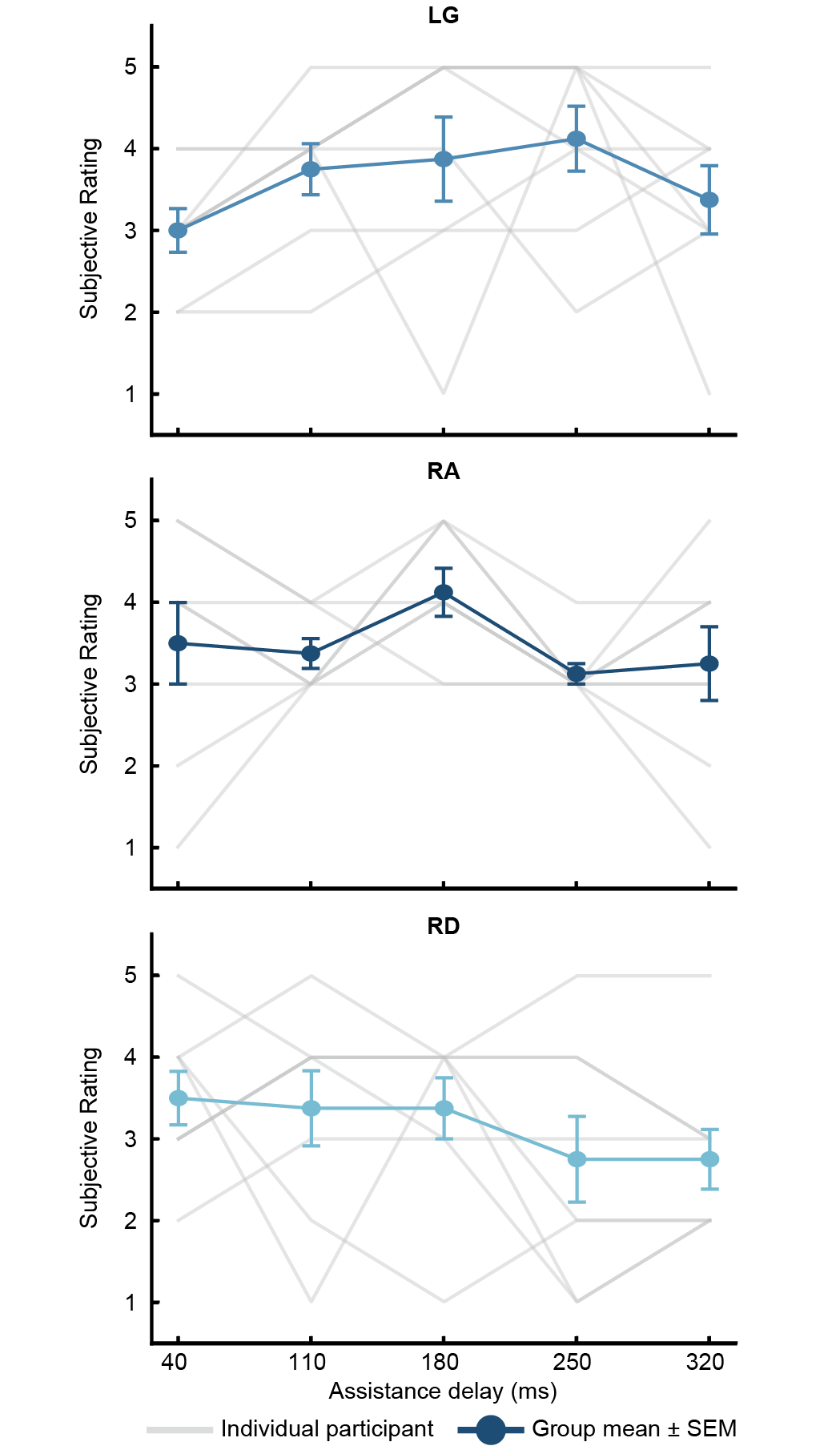}
    \caption{
       \textbf{Subjective ratings across assistance delays and locomotor tasks.} After each active-delay trial, participants rated the assistance on a five-point preference scale, considering both perceived comfort and assistance benefit, with higher scores indicating more favorable assistance. Panels, from top to bottom, correspond to level-ground (LG), ramp-ascent (RA), and ramp-descent (RD) walking. Gray lines connect ratings from the same participant across delays within each task. Colored markers and lines show the group means, and error bars indicate \(\pm\) SEM across eight participants.
    }
    \label{SubjectPreferences}
\end{figure}

    
\section{Results}
\subsection{Metabolic and Biomechanical Outcomes}
Averaged across tasks and active-delay conditions, assistance reduced all three objective outcomes relative to NoAssi: net metabolic rate by 5.24\% (\(p=0.009\)), positive biological hip joint work by 5.86\% (\(p=0.007\)), and total lower-limb positive joint work by 1.68\% (\(p=0.013\)). The dependence of these reductions on delay differed across outcomes (Outcome \(\times\) Delay interaction, \(p<0.001\); Fig.~\ref{DelayTaskEffect}). Outcome-specific analyses showed a Delay effect for positive biological hip joint work and total lower-limb positive joint work (both \(p<0.001\)), but not for net metabolic rate (\(p=0.743\)). Post hoc comparisons showed that positive biological hip joint work was lower at 40~ms than at 110, 250, and 320~ms and higher at 320~ms than at 110, 180, and 250~ms; total lower-limb positive joint work was higher at 320~ms than at 40, 110, and 180~ms (all \(p_{\mathrm{adj}}\leq0.041\)). In the exploratory linear mixed-effects models, reductions in positive biological hip joint work and total lower-limb positive joint work decreased as delay increased (both \(p<0.001\)), whereas net metabolic rate showed no linear delay-response trend (\(p=0.577\)).

Task dependence of the delay response was not detected, although assistance benefits were greater in LG and RA. The Task \(\times\) Delay interaction was not significant in the omnibus analysis (\(p=0.163\)) or in any outcome-specific analysis (all \(p\geq0.101\)), nor was the Outcome \(\times\) Task \(\times\) Delay interaction (\(p=0.713\)). The exploratory linear mixed-effects models likewise showed no task-dependent delay slopes (all \(p\geq0.222\)). Averaged across the active-delay conditions, net metabolic rate decreased by 9.75\% during LG and 7.20\% during RA but increased by 1.23\% during RD. Total lower-limb positive joint work differed across tasks (\(p=0.009\)), and Bonferroni-corrected post hoc comparisons showed a greater NoAssi-relative reduction during LG than during RD (\(p_{\mathrm{adj}}=0.002\)).

Bonferroni-corrected task-wise comparisons with NoAssi identified at least one delay that reduced each outcome during LG and RA, but none during RD. Net metabolic rate was reduced at 110, 180, 250, and 320~ms during LG and at 40, 110, and 180~ms during RA. The largest observed metabolic reductions occurred at 180~ms during LG (11.33\%) and at 110~ms during RA (8.97\%). Positive biological hip joint work was reduced at 40, 110, and 180~ms during both LG and RA, whereas total lower-limb positive joint work was reduced at 40, 110, and 180~ms during LG and at 40~ms during RA. The largest observed reductions in both positive joint-work outcomes occurred at 40~ms during LG and RA.

\subsection{Subjective Outcomes}
The highest mean subjective ratings occurred at 250~ms during LG, 180~ms during RA, and 40~ms during RD (Fig.~\ref{SubjectPreferences}). 
Ratings varied substantially across participants, and no effect of Task (\(p=0.234\)), Delay (\(p=0.126\)), or their interaction (\(p=0.295\)) was detected. Thus, subjective ratings did not identify a consistent group-level preference for a particular assistance delay in the present sample.

\section{Discussion}
Biomechanical and metabolic outcomes capture different consequences of assistance timing and may therefore favor different controller parameters.
Across tasks, reductions in positive biological hip joint work and total lower-limb positive joint work generally decreased as delay increased, whereas net metabolic rate remained relatively insensitive to delay (Fig.~\ref{DelayTaskEffect}).
Consequently, the delay associated with the greatest mechanical unloading did not necessarily provide the greatest metabolic benefit.

The contrasting delay sensitivities of positive joint work and net metabolic rate reflect the distinct physiological processes underlying these outcomes. 
Positive biological joint work provides a relatively direct measure of the mechanical response to exoskeleton torque. Increasing assistance delay shifts the applied torque relative to the biological hip moment and joint angular velocity, potentially reducing the amount of biological work displaced by the device. 
In contrast, net metabolic rate reflects the combined energetic effects of muscle activation, force production, stabilization, and compensatory responses across the body. 
Therefore, mechanical unloading at the assisted joint may be offset by altered muscular demands elsewhere.
The sensitivity of net metabolic rate to assistance delay is consistent with findings from Molinaro et al., who reported a maximum metabolic difference of approximately 2.9\% across delays ranging from 75 to 175~ms during LG walking \cite{Molinaro_Sci_Robot}. 
Despite evaluating a substantially wider delay range of 40 to 320~ms, our study shows a similar result of marginal differences in metabolic benefit across varying delay conditions. 

These outcome-dependent responses also have implications for the optimization of exoskeleton control parameters, in which prior studies have used metabolic cost as the primary objective \cite{JuajuanHILO, Stair, StrokeHip, ConorWalshHILO}.
Our results show that the delay that minimizes metabolic rate may not maximize reductions in positive biological joint work.
Controller parameters should therefore be selected according to the intended application.
When both energetic assistance and mechanical unloading are desired, metabolic and biomechanical outcomes may need to be considered jointly.

Beyond the outcome-dependent effects, the task-wise pattern suggests a potential boundary to the generalizability of the fixed state-to-torque mapping. Multiple delay conditions were associated with metabolic and mechanical benefits during LG and RA. In contrast, no individual delay was associated with a statistically reliable reduction in any evaluated outcome during RD.
By extending prior evaluations of biological torque assistance beyond LG and RA \cite{Molinaro_Sci_Robot}, the present study highlights RD as a condition in which the fixed mapping warrants further evaluation.

The mechanical demands of RD may explain the limited effectiveness of the predominantly positive-power assistance strategy. 
Positive mechanical power from an exoskeleton can offset part of the positive biological work required for propulsion during LG and for body elevation during RA \cite{UpDownhillOpt}. 
In contrast, RD requires substantially greater mechanical energy absorption and negative work to control the descent of the body \cite{UphillDownhillJointWork, HipKneeAnkleUphillDownhill}.
The current state-to-torque mapping may therefore have been poorly matched to the eccentric and braking demands of RD. 
Positive power applied during descent could also require compensatory energy absorption by the user, limiting both metabolic and mechanical benefits. 
These mechanics suggest that improving assistance during descent may require changes in the power-delivery characteristics of the mapping beyond delay alone, including negative-power or energy-absorption assistance.

The absence of statistically distinct task-dependent delay responses should not be interpreted as evidence that the locomotor tasks were biomechanically equivalent or that the fixed mapping was optimal for every task.
Although Powell et al. optimized multiple mapping parameters and the present study varied delay alone, their findings are broadly consistent with ours. 
They reported metabolic reductions of 8.6\% and 8.4\% from task-specific and task-agnostic assistance, respectively, across level walking at two speeds and 5$\degree$ incline walking, without evidence of an additional benefit from task-specific assistance \cite{Justine}. 
One possible explanation for this convergence is that biological moment-based control still generates task-dependent assistance, even when the downstream mapping remains fixed, because the estimated hip moment used as its input changes with the locomotor condition.
Within LG and RA, this input-level adaptation may have captured much of the task-related variation in hip mechanics, leaving limited additional benefit from task-specific adjustment of delay. 
The interaction results do not contradict the limited RD benefit because task-dependent delay sensitivity and the overall effectiveness of the tested mapping address distinct questions.
The present findings did not provide evidence for task-specific delay tuning within the tested conditions, but they motivate examining whether other mapping components should be adapted according to task mechanics.

More broadly, these findings suggest that the need for a task-specific state-to-torque mapping may depend less on the nominal locomotor mode than on how similarly different tasks load the assisted joint mechanically.
This criterion is inherently joint-specific because the same pair of tasks can impose similar mechanics at one joint but distinct mechanics at another.
Across graded walking, hip mechanics are modulated predominantly through positive work, whereas knee mechanics are modulated more strongly through negative work \cite{UphillDownhillJointWork, HipKneeAnkleUphillDownhill}.
Thus, a mapping that generalizes across LG and RA at the hip may not necessarily generalize across the same task set at the knee.
The existence of both shared and distinct biomechanical task spaces is supported by An et al., who identified redundancy and distinct clustering among cyclic and non-cyclic activities in a reduced biomechanical feature space \cite{myICORR}. Their analysis, however, addressed estimator training rather than mapping optimization.
A shared mapping may be more likely to generalize when tasks occupy overlapping regions of the assisted joint's mechanical state space and less likely when those regions are mechanically distinct.
Cyclicity alone cannot determine the need for task-specific mapping, as RD is cyclic yet differs substantially from LG and RA in its energy-absorption demands. 
This framework predicts that differences between shared and task-specific mappings will become more apparent across task pairs that occupy mechanically distinct regions of the assisted joint's state space.

Subjective preference should be treated as a complementary human-centered outcome rather than as a direct surrogate for metabolic or biomechanical benefit.
The lack of a consistent group-level delay preference is consistent with previous evidence that preferred assistance timing does not necessarily correspond to the timing that minimizes metabolic cost \cite{YoungTiming}.
Medrano et al.\ reported an average just-noticeable difference in metabolic rate of \(22.7\pm17.0\%\) at a 75\% correct discrimination threshold \cite{RouseMetabolicPreference}.
The metabolic differences among the active delay conditions in the present study were smaller than this average perceptual threshold, which may have limited participants' ability to discriminate among conditions.
Subjective ratings may also reflect perceived comfort and synchronization with the device in addition to objective assistance performance.

Several limitations constrain the interpretation of the present findings.
First, the sample size of eight participants limited sensitivity to task-dependent and subjective effects.
Post hoc power for the categorical task-dependent and subjective effects ranged from 23.4\% to 50.6\%. 
Conditional on the observed effect patterns and covariance structure, approximately 15 to 30 participants would have been required to achieve 80\% power.
Thus, these nonsignificant results should not be interpreted as evidence of equivalent delay-response profiles or an absence of systematic preference.
This limitation primarily concerns the null findings and does not negate the detected overall assistance effects or the Outcome \(\times\) Delay interaction.
Second, the study varied delay at 70-ms intervals while holding the other components of the state-to-torque mapping fixed.
The findings therefore neither identify a precisely optimal delay nor determine whether task-dependent benefits could emerge through changes in scaling, shaping, or other mapping components.
Instead, the conclusions are specific to delay tuning within the tested mapping and delay range.
Finally, the biomechanical analyses focused on positive joint work and did not quantify changes in biological negative work or energy absorption.
The proposed mismatch between the predominantly positive-power mapping and the mechanical demands of RD should therefore be interpreted as a biomechanical explanation rather than a directly tested mechanism. This limitation does not alter the observed task-level pattern in the measured outcomes.
Future studies should evaluate the complete state-to-torque mapping, including scaling, timing, and shaping, in larger samples while directly quantifying positive and negative biological and exoskeleton work.
Extending these evaluations across assisted joints and mechanically distinct cyclic and non-cyclic activities could clarify whether mapping generalizability is better predicted by joint-level mechanical similarity than by nominal task labels.

\section{Conclusion}
This study systematically evaluated the effects of assistance delay on the benefits provided by a biological torque controller using a robotic hip exoskeleton. 
The influence of assistance delay depended strongly on the selected outcome: shorter delays produced greater mechanical unloading, whereas metabolic benefits remained comparatively insensitive to delay.
Assistance benefits were concentrated in LG and RA and did not consistently extend to RD, suggesting that varying delay within the tested range was insufficient to overcome a possible mismatch between the predominantly positive-power mapping and the energy-absorption demands of descending locomotion.
These findings suggest that state-to-torque mappings should be evaluated according to the intended outcome and the mechanical role of the assisted joint rather than locomotor labels alone.
Shared mappings may generalize across tasks with overlapping joint-level mechanics, whereas mechanically distinct conditions may require modifications beyond assistance delay.


\bibliographystyle{ieeetr}
\bibliography{References}

\end{document}